\documentclass{INTERSPEECH2023}

\usepackage{amsmath,graphicx}

\usepackage{float}

\usepackage{hyperref}
\usepackage{cleveref}

\usepackage{booktabs}
\usepackage[table,xcdraw]{xcolor}

\usepackage{amssymb}
\usepackage{balance}

\newcommand{\eg}{e.\,g.\,}

\interspeechcameraready

\title{Bayesian Networks for the robust and unbiased prediction of depression and its symptoms utilizing speech and multimodal data}

\name{Salvatore Fara$^1$, Orlaith Hickey$^1$, Alexandra Georgescu$^{1,2}$, Stefano Goria $^1$, Emilia Molimpakis$^1$, Nicholas Cummins$^{1,2}$}
\address{
  $^1$Thymia, London, UK\\
  $^2$Institute of Psychiatry, Psychology \& Neuroscience (IoPPN), King’s College London, London, UK}
\email{\{salvatore,orlaith,alexandra,stefano,emilia,nick\}@thymia.ai, nick.cummins@kcl.ac.uk}

\begin{document}

\maketitle
 
\begin{abstract}
Predicting the presence of major depressive disorder (MDD) using speech is highly non-trivial. The heterogeneous clinical profile of MDD means that any given speech pattern may be associated with a unique combination of depressive symptoms. Conventional discriminative machine learning models may lack the complexity to robustly model this heterogeneity. Bayesian networks, however, are well-suited to such a scenario. They provide further advantages over standard discriminative modeling by offering the possibility to (i) fuse with other data streams; (ii) incorporate expert opinion into the models; (iii) generate explainable model predictions, inform about the uncertainty of predictions, and (iv) handle missing data. In this study, we apply a Bayesian framework to capture the relationships between depression, depression symptoms, and features derived from speech, facial expression and cognitive game data. Presented results also highlight our model is not subject to demographic biases. 
\end{abstract}

\noindent
\textbf{Index Terms}: Depression, Bayesian Networks, Fusion, Knowledge Integration, Missing Data, Fairness

\section{Introduction}

The healthcare sector is in urgent need of better tools to tackle the challenges of major depressive disorder (MDD) efficiently and effectively. Depression assessments are still based on self-report questionnaires which are prone to bias \cite{hunt2003self} and where the variability between individuals' interpretation of questionnaire items is high \cite{vanheule2008factor}. Furthermore, clinical interviews and observation are naturally influenced by the clinician's experience and acumen \cite{markova2009epistemology}. Collectively, this means identifying the correct diagnosis and treatment can take many years, with some studies finding untreated depression rates as high as 77\% \cite{strawbridge2022care}. There is an immediate need for a clinical decision support tool offering objective depression metrics, as easily accessible and reliably trackable as physical health ones (e.g. blood test markers). Advances in digital health and phenotyping technologies are therefore being considered integral to improving MDD-associated clinical pathways \cite{strawbridge2022care}. 

In recent years, there has been an acceleration in the number of papers centred around the application of \emph{machine learning} in the domain of digital health. These works include analyses of speech, facial expressions and cognitive assessments to provide objective measurement criteria to aid in MDD diagnosis \cite{cohn2018multimodal, girard2015automated}. A potential shortcoming of such works, however, is that they have almost exclusively been focused on supervised modelling paradigms learning how to partition data based on subjective depression scales, such as the 8-item Patient Health Questionnaire (PHQ-8; \cite{kroenke2009phq}), thereby also becoming subject to the same concerns around self-report subjectivity. Moreover, they mainly utilise large multivariate feature spaces and deep learning models which lack transparency regarding how their decisions are being made \cite{von2021transparency}. Alongside this lack of explainability, such approaches also lack the ability to incorporate expert opinion into the model and are unable to handle missing data robustly. 

Bayesian Networks (BN) offer a natural framework to satisfy all the above requirements, which are common in healthcare modelling. Indeed, a few recent works have successfully adopted BNs to tackle mental health modelling problems; for example, \cite{MCLACHLAN2020101912, kyrimi2021comprehensive}. However, the predictors in these approaches have been simple demographics, biological or environmental factors, as opposed to rich multimodal datasets that can include audio and video data. Only a very small number of works have explored the use of BNs for detecting depression from speech \eg \cite{Yang2019_MDD, espinola2021detection}. These works, however, are focused on the classification of depression severity; they do not consider joint classification with symptoms or the inclusion and effects of confounding factors. 

In this study we propose a novel BN model for joint MDD and depression symptoms classification given a multimodal feature set containing speech, facial expression and cognitive game data gathered at thymia \cite{fara2022speech}. 
We then present a range of experiments demonstrating the model's performance under different realistic use case scenarios, including varying degrees of missing data and integration of expert knowledge. 

The main novel contribution of this work is a methodology for incorporating speech and video data in a BN model that achieves strong performance in MDD classification. We also highlight its potential as a clinical decision-support tool by providing results for individual core MDD symptoms and give a detailed breakdown of performance according to key sociodemographic factors.

\section{Experimental Corpora}
\label{sec:typestyle}
This section describes the collection and preprocessing of the data used in our expirements. 

% our key experimental settings including %the thymia 
% our dataset and introduces our Bayesian network.

\subsection{Dataset}
\label{ssec:pagestyle}

We trained and validated our models on our data collected in-house. This data collection received ethical approval from the Association of Research Managers and Administrators. Publicly available speech-depression datasets such as the Audio-Visual Depressive (AViD) corpus \cite{Valstar13-A2T} and Distress Analysis Interview Corpus (DAIC) \cite{gratch_distress_2014} do not have the required meta-data to support our stated analytical aims. Further, these data have been in the public domain for 10 years so subject to concerns relating to overfitting and multiple hypothesis testing.

% Dataset statistics table (with Device).
\begin{table*}[t!]
  \caption{Sociodemographic, depression and activity distributions in the experimental data.}
  \vspace{-2mm}
  \label{tab:demographics}
  \centering
  \renewcommand{\arraystretch}{1.1}
  \begin{tabular}{c c c c c c c c c c c c}
    \toprule
    \multicolumn{1}{c}{\textbf{Group}} &
    \multicolumn{2}{c}{\textbf{Gender}} &
    \multicolumn{2}{c}{\textbf{Age}} &
    \multicolumn{2}{c}{\textbf{Country}} &
    \multicolumn{2}{c}{\textbf{Device}} &
    \multicolumn{3}{c}{\textbf{Cumulative Activity Time}} \\
    \multicolumn{1}{c}{\textbf{}} &
    \multicolumn{1}{c}{\textbf{Male}} &
    \multicolumn{1}{c}{\textbf{Female}} &
    \multicolumn{1}{c}{\textbf{$<$ 36}} &
    \multicolumn{1}{c}{\textbf{$\geq$ 36}} &
    \multicolumn{1}{c}{\textbf{UK}} &
    \multicolumn{1}{c}{\textbf{US}} &
    \multicolumn{1}{c}{\textbf{Phone}} &
    \multicolumn{1}{c}{\textbf{PC}} &
    \multicolumn{1}{c}{\textbf{Paragraph}} & 
    \multicolumn{1}{c}{\textbf{Image}} &
    \multicolumn{1}{c}{\textbf{n-Back}} \\
    \midrule
    PHQ-8 $\geq$ 10 & 135 & 220 & 215 & 140 & 202 & 153 & 19 & 336 & 4:09:55 & 5:23:22 & "1 day, 19:15:27" \\ 
    PHQ-8 $<$ 10 & 496 & 485 & 574 & 407 & 424 & 557 & 42 & 938 & 14:15:52 & 19:10:56 & "5 days, 21:20:47" \\ 
    \bottomrule
  \end{tabular}
  \vspace{-5mm}
\end{table*}

The experimental data used in this study consist of 1,336 English-speaking participants who performed a series of short online activities within a single session on %the thymia 
our Research Platform \cite{fara2022speech} \footnote{The thymia Research Platform allows the hosting of complex, remote, multimodal studies on a smart device. During various activities (e.g. questionnaires, cognitive games etc.), data from the device’s camera, keyboard, mouse/trackpad and/or touch screen can be streamed to a secure backend. The platform is fully HIPAA-compliant, 2018 EU GDPR-compliant, is ISO27001-certified and NHS Toolkit-compliant.} using their own personal devices (Table~\ref{tab:demographics}). We previously presented a portion of this dataset with a smaller number of participants and a focus on audio (acoustic, prosodic and linguistic) and cognitive data gathered from two thymia activities, i.e. an Image Description Task and the n-Back Task \cite{nikolin2021investigation}, as well as individual PHQ-8 items; see \cite{fara2022speech}. In the present study, we expand the number of data modalities to include video data recorded during the Image Description Task, as well as additional audio data gathered during a Paragraph Reading Task. Additionally, we include information on the type of personal device that was used to perform the activities.  

\subsection{Data Availability} 
Due to licensing and IP considerations, we are not at this moment making our dataset generally publicly available. However, we are open to partnering with research institutes and individual academics including data sharing upon request.

%\subsection{Thymia Activities}
\subsection{Data Collection Activities}
\label{ssec:activities}

We focus on data gathered through three %thymia activities: 
data collection activities: the n-Back Task, the Image Description Task and the Paragraph Reading Task. The first two activities have been previously described in detail in \cite{fara2022speech}. The Paragraph Reading Task required participants to read aloud a short story (Aesop fable ``The North Wind and the Sun'' widely used within phonetics \cite{international1999handbook}) while their voice was being recorded via their device's microphone. Herein we abbreviate the activity names to ``\textit{n-Back}'', ``\textit{Image}'', and ``\textit{Paragraph}''. 

\subsection{Data Selection}

A total of 1,898 participants enrolled to the study. Participants were excluded from the dataset if either of these applied: (1) they did not complete all three data collection activities; (2) their recordings were corrupted by technical problems (camera/mic malfunctioning); (3) they did not comply to the tasks (did not speak in the speech tasks). On the basis of these criteria, 1,336 participants were selected. 

\subsection{Data Preprocessing}

Audio recordings from the speech eliciting activities were converted to single-channel wave files at 16kHz sampling rate using FFmpeg software. Speech tokens were then extracted from the audio files using Amazon Web Services (AWS) Speech-To-Text service Amazon Transcribe.

\subsection{Features}
\label{ssec:features}
 
Data from the three %thymia 
activities was processed to extract a total of 322 features which included: 8 cognitive features (n-Back), 97 video features (Image), 88 extended Geneva Minimalistic Acoustic Parameter Set (eGeMAPS) acoustic and prosodic features \cite{eyben2015geneva} (from both Image and Paragraph), 24 linguistic features (Image), as well as an additional curated set of 17 fine-grained acoustic features (Paragraph). Details on the cognitive, eGeMAPS and speech features were previously provided and can be found in \cite{fara2022speech}.   

The video features were extracted using Visage Technologies Software. The software extracted features related to facial translation, rotation and gaze in the 3D space, action units and emotions.  An estimated face scale was also provided to normalise values, allowing for changes in a participant's distance from the screen and camera. 

The fine-grained acoustic features consist of summary statistics of the formant trajectories extracted from specific voiced audio segments of the Paragraph audio recordings. We used audio segments corresponding to three sets of words chosen to isolate the following vowel sounds \cite{scherer2015self}: /i/ (`wind', `which', `he', `his', `him'), /u/ (`should', `could', `took', `two'), /a/ (`hard', `last', `and', `at').

\section{Bayesian Network Model}
\label{ssec:propmodel}

\subsection{Model Definition}

\begin{figure}[t]
  \centering
  \includegraphics[width=0.95\linewidth]{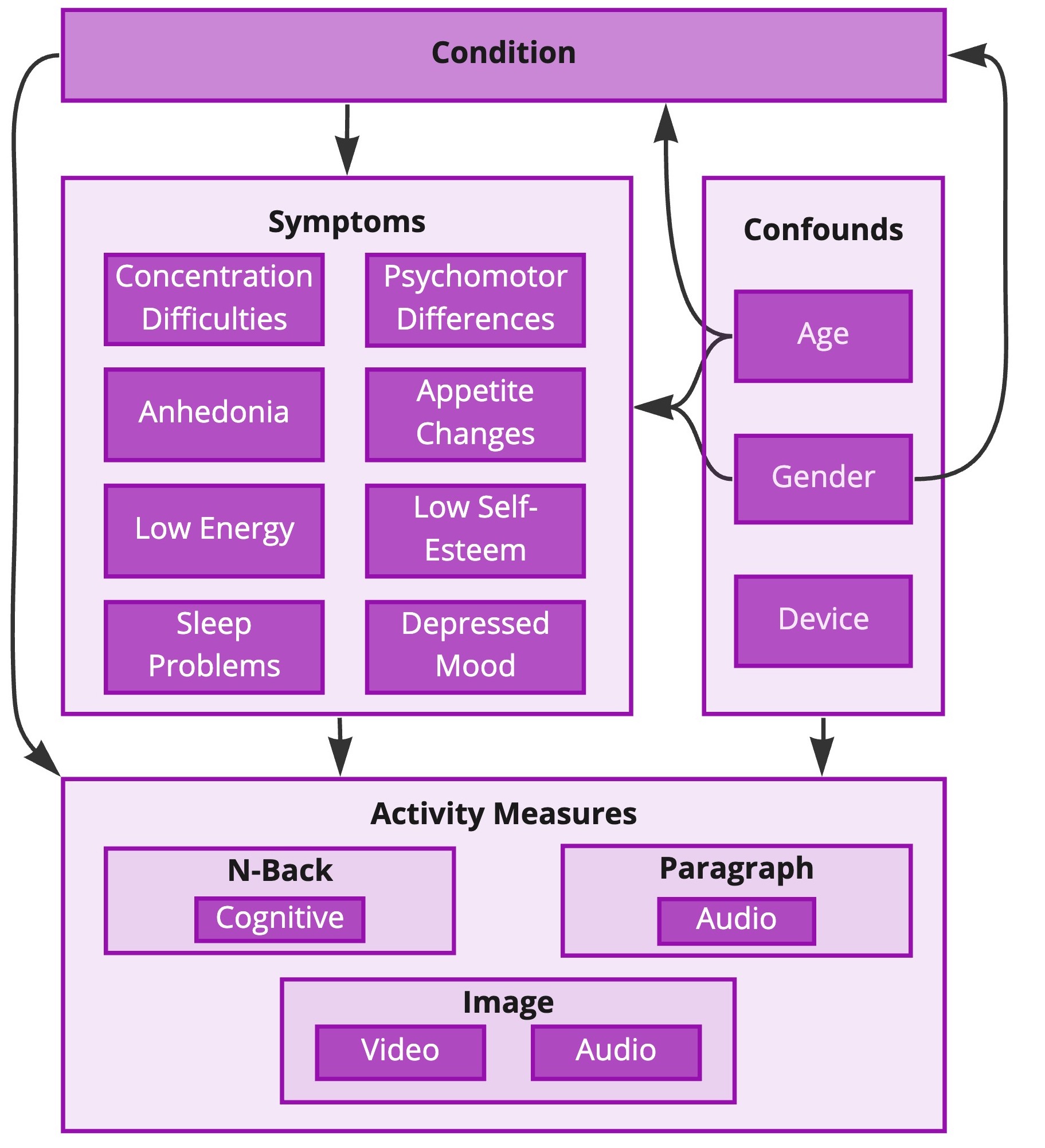}
  \caption{%Snippet of the 
  Overview of the proposed multimodal Bayesian network model.}
  \label{fig:model}
  \vspace{-5mm}
\end{figure}

Bayesian Networks (BNs) are probabilistic graphical models that specify the joint distribution by defining a set of conditional independence rules that can be easily mapped to a Directed Acyclical Graph (DAG) \cite{10.5555/1795555}. Our BN model is composed of four groups of variables: \textsc{confounds}, \textsc{condition}, \textsc{symptoms} and \textsc{activity} measures (Figure~\ref{fig:model}).

The \textsc{confounds} group includes age and gender as demographic variables,  %two variables describing the participant's demographics, i.e. age and gender, 
and a third variable indicating the type of personal device used by the participant to perform the session on %the thymia 
our Research Platform. The age variable is modelled as a categorical distribution with four categories representing four age groups (i.e. 18-25, 26-35, 36-45, 46-100), while both gender and device are modelled as Bernoulli distributions with categories `male'/`female' and `smartphone'/`PC' respectively. In the following we use $A \in \{0,1,2,3\}$, $G \in \{0,1\}$ and $D \in \{0,1\}$ to indicate the age group, gender and device respectively.  

The \textsc{condition} variable indicates the presence (PHQ-8 $\geq$ 10) or absence (PHQ-8 $<$ 10) of depression. To capture the variation of depression incidence across age groups and genders, we model the condition $C$ as a Bernoulli distribution: 
\begin{equation} \label{eq:condition}
p(C|A, G) = Ber(C|logistic(f_{c}(A,G)))
\end{equation}
where  
\begin{equation} \label{eq:f_c}
f_{c}(A,G) = \omega_{c,0} + \omega_{c,a}A + \omega_{c,g}G .
\end{equation}

The \textsc{symptom} variables represent the individual PHQ-8 items. In order to simplify the model, each symptom is converted from its original 4-point scale to a binary scale indicating `low' and `high' symptom levels. The symptom-specific binarisation thresholds are calculated from a logistic regression of each individual symptom on the condition variable. In order to capture inter-symptom conditional dependencies, the symptom variables are embedded in an inter-symptom DAG estimated using the DirectLiNGAM graph discovery algorithm \cite{zeng2022causal}. Each binary symptom variable $S_{s}$, with $s \in \{0, 1, ..., 7\}$, is modelled as a Bernoulli distribution: 
\begin{equation} \label{eq:symptom}
p(S_{s}|A, G, C, \pmb{P_{s}}) = Ber(S_{s}|logistic(f_{s}(A, G, C, \pmb{P_{s}})))
\end{equation}

where variables in bold are vectors, $\pmb{P_{s}}$ is a column vector of $k_{s}$ parent symptoms of $S_{s}$ as specified by the inter-symptom DAG and 

\begin{equation} \label{eq:f_s}
\begin{split}
f_{s}(A, G, C, \pmb{P_{s}}) = \enspace&\omega_{s,0} + \omega_{s,a}A + \omega_{s,g}G \enspace + \\ &\omega_{s,c}C + \pmb{\omega_{s, p} P_{s}}
\end{split}
\end{equation}
with $\pmb{\omega_{s, p}} \in \mathbb{R}^{k_s}$. In the following we use $\pmb{S}$ to indicate the column vector of all binary symptoms.

The \textsc{activity} measures are derived from the feature sets described in the previous section by applying two processing steps. First, standard rescaling is applied to all features individually. Second, supervised PCA \cite{carson2021augmentedpca} is applied to each feature set independently using the condition variable as target. The first two principal components of each feature set are then selected, yielding a total of 16 activity measures (2 from N-Back, 10 from Image, 4 from Paragraph). Each activity measure variable $M_{m}$, with $m \in \{0, 1, ..., 15\}$, is modelled as a Gaussian distribution 
\begin{equation} \label{eq:measure}
p(M_{m}|A, G, D, C, \pmb{S}) = \mathcal{N}(M_{m}|f_{m}(A, G, D, C, \pmb{S}), \sigma_{m}^{2})
\end{equation} 
where 
\begin{equation} \label{eq:f_m}
\begin{split}
f_{m}(A, G, D, C, \pmb{S}) =\enspace&\omega_{m,0} + \omega_{m,a}A + \omega_{m,g}G \enspace + \\
&\omega_{m,d}D + \omega_{m,c}C + \pmb{\omega_{m, s} S} 
\end{split}
\end{equation}
with $\pmb{\omega_{m, s}} \in \mathbb{R}^{8}$. In the following we use $\pmb{M}$ to indicate the column vector of all activity measures.

The full BN model describing the joint probability distribution over all the variables described above is then given by
\begin{equation} \label{eq:model}
\begin{split}
p(\pmb{M}, \pmb{S}, C, A, G, D) =\enspace &\prod_{m=0}^{15}{p(M_{m}|A, G, D, C, \pmb{S})} \times\\
&\prod_{s=0}^{7}{p(S_{s}|A, G, C, \pmb{P_{s}})} \times \\
&p(C|A, G)\times \\
&p(A)\times p(G)\times p(D) .
\end{split}
\end{equation} 

% \begin{equation} \label{eq1}
% \begin{split}
% p(C|A, G) &= Ber(C|logistic(f_{a}(A,G))) \\
% f_{a}(A,G) &= \omega_{c,0} + \omega_{c,a}A + \omega_{c,g}G
% \end{split}
% \end{equation}
% \begin{equation} \label{eq:condition}
% p(C|A, G, \pmb{\omega_{c}}) = Ber(C|logistic(\omega_{c,0} + \omega_{c,a}A + \omega_{c,g}G))
% \end{equation}

% The age variable is modelled as a categorical distribution
% \begin{equation} \label{eq:age}
% p(A = i) = \theta_{A = i}
% \end{equation}
% where $i \in \{1,2,3,4\}$ represents one of four age groups (i.e. 18-25, 26-35, 36-45, 46-100) and $\theta_{A = i}$ is the frequency of age group $i$.

\subsection{Model Implementation and Training}
\label{ssec:modeltraining}

% \begin{table}[t!]
% \caption{Bayesian network model parameters.}
% \label{tab:feature-ranks}
% \centering
% \begin{tabular}{@{}lrr@{}}
% \toprule
% \textbf{Parameter} & \textbf{Symbol} & \textbf{Prior} \\ \midrule
% Age group frequencies & $\pmb{\theta_{a}} \in R^{4}$ & Dirichlet(\pmb{1}) \\ \bottomrule
% \end{tabular}
% \end{table}

We implemented the BN model using the probabilistic programming library NumPyro (version 0.11.0) \cite{phan2019composable} and Python 3.9.15. Model training was performed via the Markov Chain Monte Carlo (MCMC) inference of model parameters using the No-U-Turn sampler (NUTS) algorithm in NumPyro, with 4 Markov chains and 1000 samples per chain. We used the following prior distributions for the model parameters: $Dirichlet(K=4, \alpha=1)$ for the group frequencies of the age variable; $Beta(\alpha=1, \beta=1)$ for the Bernoulli probabilities of the gender and device variables; $\mathcal{N}(\mu=0, \sigma=1)$ for all $\omega$ parameters in Equations~(\ref{eq:f_c}), (\ref{eq:f_s}) and (\ref{eq:f_m}); $LogNormal(\mu=0, \sigma=1)$ for the $\sigma_{m}$ parameters in Equation~(\ref{eq:measure}).

\begin{figure}[t!]
  \centering
  \includegraphics[width=0.95\linewidth]{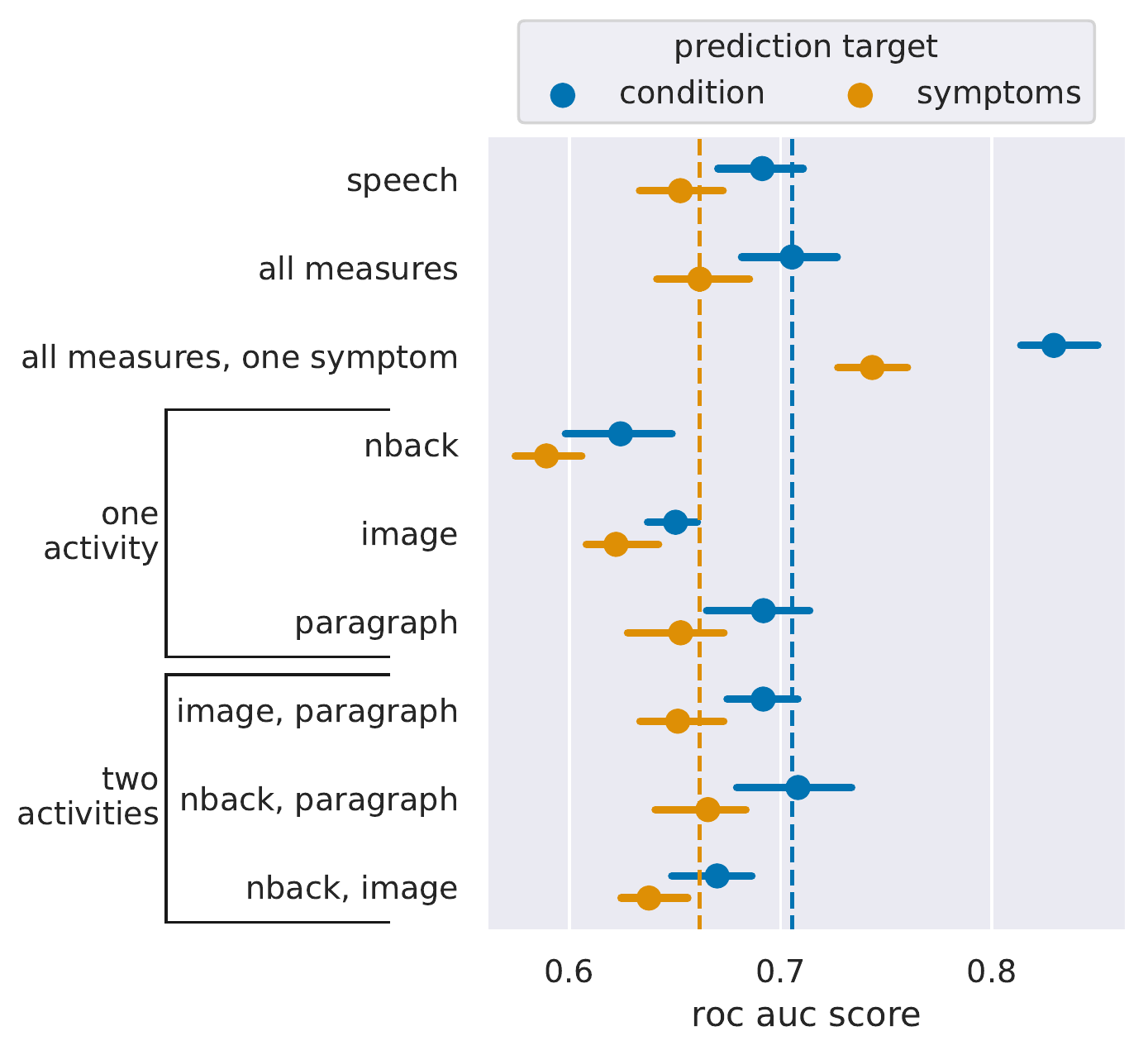}
  \caption{
  BN model performance in the joint prediction of condition and symptoms given the observation of other sets of variables. Averages (with 95\% CI) across 5 cross-validation test folds are shown. Dashed lines highlight model performances when all measures are observed.  
  }
  \vspace{-5mm}
  \label{fig:cv_results}
\end{figure}

\subsection{Model Evaluation}

We performed a stratified 5-fold cross-validation to evaluate model performance. %in a range of inference tasks. %We used stratification to keep the same proportion of gender, age groups and PHQ-8 distribution across training and test sets for each cross-validation fold. 
The same proportion of gender, age groups and PHQ-8 distribution was kept across the training and test sets for each fold. We use the area under the receiver operating characteristic curve (ROC-AUC) as our evaluation metric. % report results in terms of the area under the receiver operating characteristic curve (ROC-AUC). 

\subsection{Benchmark model}

In order to benchmark the BN model, we additionally trained a multi-output Random Forest classifier (RF) which received the same 16 input activity measures used by the BN model. The model had 500 trees with maximum depth of 5 and was implemented using the scikit-learn package (version 1.0) \cite{scikit-learn}. The maximum depth was set via grid search over [1, 5, 10, 15, 20] using a 5-fold cross-validation.  

% \subsection{Code Availability}
% The source code required for conducting experiments will be made publicly available upon publication of the paper under a dual license that allows free usage for research purposes.

\section{Results and Discussions}
\label{sec:majhead}

% \begin{figure}[t]
%   \centering
%   \includegraphics[width=0.95\linewidth]{figures/weight_params.pdf}
%   \caption{%Snippet of the 
%   Weight parameters with greatest effect size. }
%   \label{fig:model_example}
%   \vspace{-5mm}
% \end{figure}

% Results table
\begin{table}[t!]
    \caption{Mean and SD AUC of our BN and RF models. For BN we add results when slicing the data for gender, age and country.  
  }
%   \vspace{-2mm}
  \label{tab:results}
  \centering
  \begin{tabular}{c c c c c c}
    \toprule
    \multicolumn{1}{c}{\textbf{Model}} &
    \multicolumn{1}{c}{\textbf{Population}} &
    \multicolumn{4}{c}{\textbf{Target}} \\
    ~ & ~ & 
    \multicolumn{2}{c}{\textbf{Condition}} &
    \multicolumn{2}{c}{\textbf{Symptoms}} \\
    ~ & ~ & 
    \textbf{Mean} & \textbf{SD} &
    \textbf{Mean} & \textbf{SD} \\
    \midrule
    BN & Overall & 0.705 & 0.029 & 0.662 & 0.026 \\ 
    ~ & Female & 0.700 & 0.055 & 0.668 & 0.039 \\ 
    ~ & Male & 0.705 & 0.042 & 0.648 & 0.027 \\ 
    ~ & UK & 0.660 & 0.075 & 0.613 & 0.050 \\ 
    ~ & US & 0.746 & 0.019 & 0.707 & 0.027 \\ 
    ~ & Age $<$ 36 & 0.693 & 0.036 & 0.644 & 0.034 \\
    ~ & Age $\geq$ 36 & 0.726 & 0.034 & 0.687 & 0.017 \\
    \midrule
    RF & Overall & 0.714 & 0.027 & 0.665 & 0.028 \\  
    \bottomrule
  \end{tabular}
  \vspace{-5mm}
\end{table}

Given the generative nature of a BN model, any of its variables or groups of variables can be chosen as targets in a prediction task. Of particular interest is the task of predicting \textsc{condition} and \textsc{symptoms} given the observation of other variables in the model. We performed a set of experiments to evaluate the model performance in this joint prediction task under several realistic scenarios (Figure~\ref{fig:cv_results}). Overall, the experiments showed an increase in predictive performance with the amount of observed variables in the model, and a generally higher performance for \textsc{condition} than \textsc{symptoms}. %The lowest performance was found when the \textsc{confounds}, age and gender, are the only available information, with an average ROC-AUC of 0.58 for \textsc{condition} and 0.55 for the \textsc{symptoms}.
When all \textsc{activity} measures are available, the average ROC-AUC is above 0.7 for \textsc{condition} and 0.66 for \textsc{symptoms}. 

Additionally, we evaluated the performance when only subsets of \textsc{activity} measures are available as input to the model (Figure~\ref{fig:cv_results}). These experiments correspond to common real-life scenarios in which a participant does not perform the full set of activities or opts out of %audio/video 
recording. This set of experiments revealed an increase in model performance as more activities are observed, with paragraph measures having the strongest positive impact.  

Finally, we performed a set of experiments where all \textsc{activity} measures plus one \textsc{symptom} are observed (Figure~\ref{fig:cv_results}). This simulates the scenario in which reliable information about the presence or absence of a symptom is available to the clinician using the model. In this scenario, we observed that the predictive performance further improves for both \textsc{condition} and the other unknown \textsc{symptoms}. 

\begin{figure}[t]
  \centering
  \includegraphics[width=0.9\linewidth]{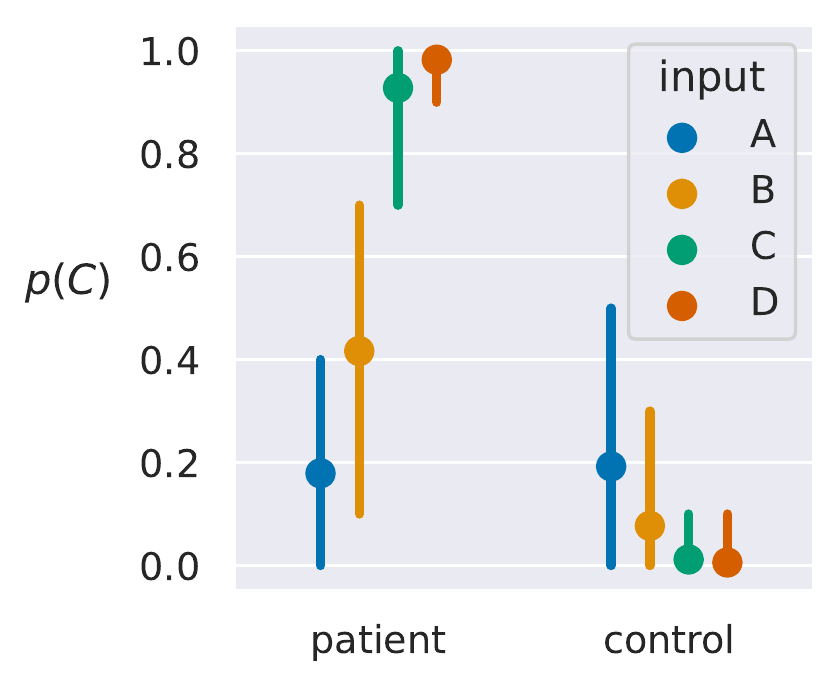}
  \caption{
  Raw model predictions of condition probability in two sample participants, one patient and one control, for four sets of inputs: A = confounds, B = confounds + n-back, C = confounds + n-back + paragraph, D = confounds + n-back + paragraph + sleep symptom. Error bars denote 95\% credible intervals.}
  \vspace{-5mm}
  \label{fig:model_example}
\end{figure}

To assess the robustness of the model %under
for potential demographics biases, we also performed a segmentation of the performance metrics across gender, age and country (Table~\ref{tab:results}). The small differences across the demographics splits we considered suggest that our results are not biased. 

When benchmarking our BN model against a RF classifier, we can see that both have similar performances (Table~\ref{tab:results}). This could be viewed as a limitation of our model; however, it is worth considering that we needed only a single BN to perform multiple predictions, whereas separate RF models need to be trained for each inference task. Furthermore, although the
performance of our BN model is below that of other approaches in the literature, we believe that our results are a true reflection of what could be expected from a dataset of this size once biases have been minimised \cite{berisha22_interspeech}.

Finally, the main purpose of our BN model is to serve as a support tool for clinical decision-making. The model allows for (i) the integration of multimodal information alongside speech; and (ii) the clinician using it to provide their expert knowledge.  % expert knowledge provided by the clinician using it. 
Given its ability to generate predictions despite missing information, this model naturally lends itself to be used as part of an iterative screening process. For example, the clinician may decide to administer only a subset of activities to a patient, 
then consult the model predictions and decide whether other information may be needed to support a diagnosis, subsequently administering additional activities, asking the patient about their sleep patterns or investigating other symptoms (Figure~\ref{fig:model_example}).

\section{Conclusions}
\label{sec:conc}
This work represents a proof-of-concept for validating a BN model, demonstrating its performance as a robust speech-based MDD prediction tool. We also highlight the potential of this model under different real-world operating conditions and assess it for potential biases in core sociodemographic factors. A limitation of the current model is the reduced set of confounding variables. Future research will explore a larger set of confounds, such as life events that could affect mood (e.g. loss or change of jobs) or health problems that could affect voice (e.g. having a cold). An additional limitation of the current model is the lack of time dynamics, which limits its scope to static one-off predictions. 

In future work, we aim to collect a longitudinal dataset and to expand the model to a dynamic BN, in order to enable its application to other clinical reasoning tasks where time is a critical factor, such as prognosis. Further steps will be to explore and document model interpretability, as well as to investigate specificity to MDD by studying control datasets of e.g. bipolar disorder or adjustment disorder. 

% \section{Acknowledgements}
% \textit{Blind for review} \textcolor{white}{ This paper also represents independent research part funded by the National Institute for Health Research (NIHR) Maudsley Biomedical Research Centre at South London and Maudsley NHS Foundation Trust and King’s College London. The views expressed are those of the author(s) and not necessarily those of the NHS, the NIHR or the Department of Health and Social Care.}

% \ifinterspeechfinal
%      The INTERSPEECH 2023 organisers
% \else
%      The authors
% \fi
% would like to thank ISCA and the organising committees of past INTERSPEECH conferences for their help and for kindly providing the previous version of this template.

% As a final reminder, the 5th page is reserved exclusively for references. No other content must appear on the 5th page. Appendices, if any, must be within the first 4 pages. The references may start on an earlier page, if there is space.

\balance
\bibliographystyle{IEEEtran}
% \bibliography{mybib}

\end{document}